\documentclass{article}

\usepackage{microtype}
\usepackage{graphicx}
\usepackage{subfigure}
\usepackage{booktabs} 
\usepackage{multicol,multirow}
\usepackage{bbding}
\usepackage{amsmath}
\usepackage{amssymb}
\usepackage{mathtools}
\usepackage{amsthm}
\usepackage{tcolorbox}
\definecolor{dkblue}{rgb}{0,0.08,0.45}

\theoremstyle{plain}
\newtheorem{theorem}{Theorem}[section]

\newtheorem{conjecture}[theorem]{Conjecture}

\theoremstyle{definition}
\newtheorem{definition}[theorem]{Definition}
\newtheorem{assumption}[theorem]{Assumption}
\newtheorem{example}[theorem]{Example}
\theoremstyle{remark}

\newcommand{\mE}{\mathbb{E}}

\newcommand{\mR}{\mathbb{R}}

\newcommand{\calB}{\mathcal{B}}
\newcommand{\calE}{\mathcal{E}}

\newcommand{\calM}{\mathcal{M}}

\newcommand{\calN}{\mathcal{N}}
\newcommand{\calG}{\mathcal{G}}
\newcommand{\calV}{\mathcal{V}}
\newcommand{\calU}{\mathcal{U}}

\newcommand{\pa}{\text{Pa}}
\newcommand{\de}{\text{De}}
\newcommand{\doop}{\text{do}}
\newcommand{\Dec}{\text{Dec}}

\def\given{{\,|\,}}

\usepackage{hyperref}

\usepackage[accepted]{icml2023}

\icmltitlerunning{Diffusion Model in Causal Inference with Unmeasured Confounders}

\begin{document}

\twocolumn[
\icmltitle{Diffusion Model in Causal Inference with Unmeasured Confounders}




\begin{icmlauthorlist}
\icmlauthor{Tatsuhiro Shimizu}{waseda}
\end{icmlauthorlist}

\icmlaffiliation{waseda}{School of Political Science and Economics, Waseda University, Tokyo, Japan}

\icmlkeywords{Causal Inference, Diffusion Model, Unmeasure Confounders}

\vskip 0.3in
]
\icmlcorrespondingauthor{Tatsuhiro Shimizu}{t.shimizu432@akane.waseda.jp}



\printAffiliationsAndNotice{}  

\begin{abstract}
    We study how to extend the use of the diffusion model to answer the causal question from the observational data under the existence of unmeasured confounders. In Pearl's framework of using a Directed Acyclic Graph (DAG) to capture the causal intervention, a Diffusion-based Causal Model (DCM) was proposed incorporating the diffusion model to answer the causal questions more accurately, assuming that all of the confounders are observed. However, unmeasured confounders in practice exist, which hinders DCM from being applicable. To alleviate this limitation of DCM, we propose an extended model called Backdoor Criterion based DCM (BDCM), whose idea is rooted in the Backdoor criterion to find the variables in DAG to be included in the decoding process of the diffusion model so that we can extend DCM to the case with unmeasured confounders. Synthetic data experiment demonstrates that our proposed model captures the counterfactual distribution more precisely than DCM under the unmeasured confounders. 
\end{abstract}

\section{Introduction}
Causal inference is the study of identifying the causal relationships between variables of one's interest and developing the estimator for the estimands, such as the Average Treatment Effect (ATE), from the observational data. With ATE, for instance, we can use observational data to determine the personalized medicine \cite{sanchez2022causal} that maximizes the outcome, such as recovery from a disease. There are two mainstreams in causal inference: the Potential Outcome (PO) framework \cite{imbens2015causal} and the Directed Acyclic Graph (DAG) framework \cite{pearl2016causal}. In the DAG framework, Chao et al. (2023) \cite{chao2023interventional} proposed the algorithm called the \textit{Diffusion-based Causal Model (DCM)} that allows us to sample from the target distribution of our interest, by which we can calculate the approximation of ATE, outperforming the state-of-the-art algorithms \cite{sanchez2021vaca} and \cite{khemakhem2021causal}. However, only under causal sufficiency can the DCM sample from the target distribution, which requires the complete observation of all the confounders, which often does not hold in practice where confounders are the variables that affect both the cause and outcome variables of our interest. For instance, we often cannot observe stress levels, physical activities, mental health, sleep patterns, and genetic factors. To overcome the limitation of DCM, we extend it and propose a new algorithm to be able to estimate the ATE even under the existence of the unmeasured confounders by including the nodes that satisfy the backdoor criterion \cite{pearl2016causal} in both training and sampling phases of the algorithm, which tells us which variables we should adjust. To illustrate the applicability of a new algorithm where unmeasured confounders exist, we conduct the synthetic data experiment for both simple and complex underlying data-generating processes. The experiment shows that our new algorithm samples precisely from the ground truth target distribution where DCM fails for both cases.
\begin{figure}[t]
    \includegraphics[scale=0.35]{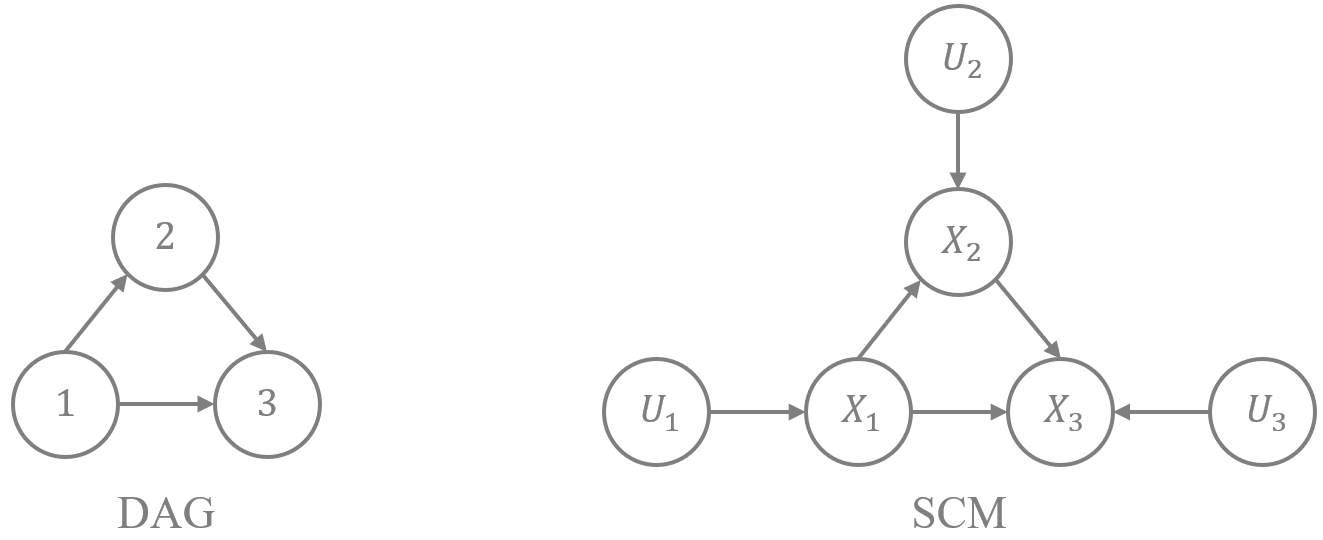}
    \centering
    \caption{DAG with three nodes and edges and the corresponding SCM with three exogenous and endogenous nodes}
    \label{fig_dag_scm}
\end{figure}
\section{Background}
We formulate the data-generating process, intervention, and causal effect in Pearl's \cite{pearl2016causal} framework. Firstly, DAG is the main element of Pearl's framework and is defined as follows.
\begin{definition}[Directed Acyclic Graph]
\label{dag}
DAG $\calG = (\calV, \calE)$ is a pair of the set of nodes $\calV$ and the set of edges $\calE$ where $\calV = \{1, \cdots, d\}$ and $\calE = \{(i, j) \colon \exists \quad \text{edge from node $i$ to $j$}\}$. DAG expresses variables by nodes and causal relationships by edges.
\end{definition}

DAG only represents the relationship between nodes on which nodes affect which nodes. To quantify how the model generates the variables in terms of distributions and functions, we introduce a \textit{structural causal model (SCM)}. We assume that we sample the observational data from the underlying SCM.

\begin{definition}[Structural Causal Model]
\label{def_SCM}
Structural Causal Model (SCM) $\calM = (\calU, \calV, f)$ is the tuple of the set of exogenous variables $\calU = \{U_1, \cdots, U_d\}$, the set of endogenous variables $\calV = \{X_1, \cdots, X_d\}$, and the set of structural equations $f=\{f_1, \cdots, f_d\}$ such that for each $i \in [d]$, the endogenous variable satisfies $X_i = f_i(\pa(X_i), U_i)$ where $\pa(X_i)$ is the set of the parent nodes of $X_i$ and $d$ is the number of endogenous or exogenous nodes.
\end{definition}

Fig. \ref{fig_dag_scm} illustrates the examples of DAG $\calG = (\calV, \calE)$ where $\calV = \{1, 2, 3\}$ and $\calE = \{(1, 2), (1, 3), (2, 3)\}$ and SCM $\calM = (\calU, \calV, f)$ with three endogenous and exogenous nodes where $\calU = \{U_1, U_2, U_3\}$, $\calV = \{X_1, X_2, X_3\}$, and $f = \{f_1, f_2, f_3\}$. 

In SCM, we assume that there exist unknown distributions of exogenous variables. We generate $n$ independent and identically distributed samples from the distribution for each exogenous variable, but we do not observe them. Then, we observe endogenous variables according to the underlying structural equation for each. Thus, the observational data is $X \in \mR^{n \times d}$ where $n$ is the number of samples and $d$ is the number of nodes

Only with the observational data $X \in \mR^{n \times d}$ are we not sure about the parent nodes of each endogenous variable. Such information can be defined by the \textit{topological order} as follows.

\begin{figure}[t]
    \includegraphics[scale=0.28]{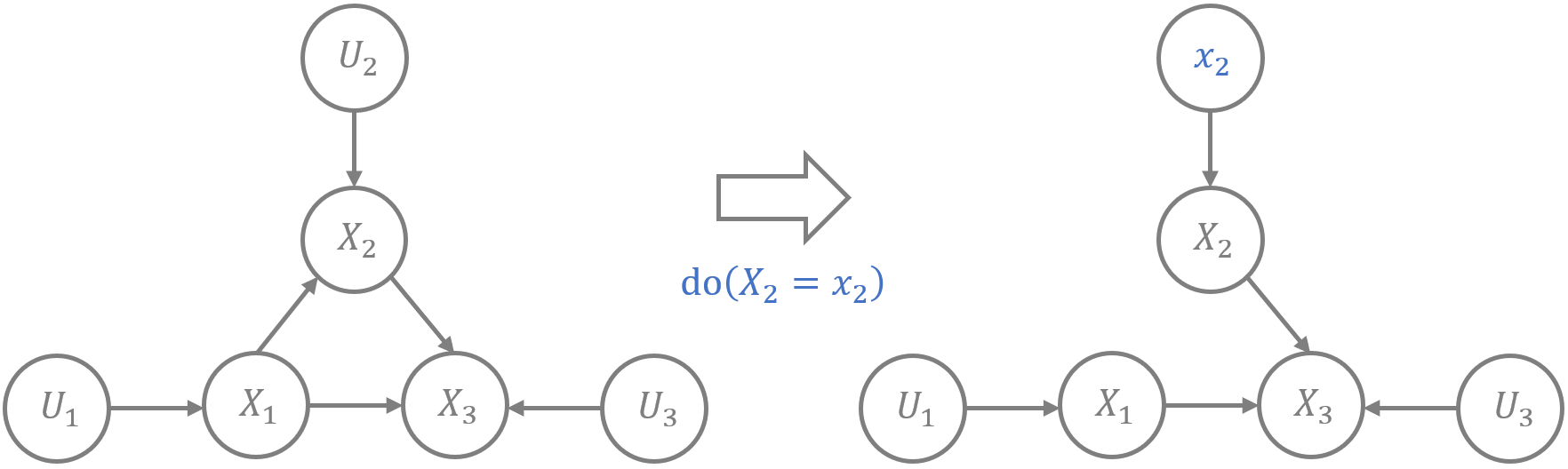}
    \centering
    \caption{do operator where we intervened in the node $X_2 = x_2$ in SCM with three exogenous and endogenous nodes}
    \label{fig_do}
\end{figure}

\begin{definition}[Topological Order]
    Topological order $\pi = (\pi_1, \cdots, \pi_d)$ is a permutation of $d$ nodes in SCM such that $\pi_i < \pi_j \iff X_j \in \de(X_i)$ for all $X_i, X_j \in \calV$ such that $i \neq j$ where $\de(X_i)$ is the set of the descendant nodes of $i$.
\end{definition}
The problem of finding the topological order given the observational data is called \textit{causal discovery} \cite{spirtes2000causation}, \cite{glymour2019review}. As this problem is computationally intensive and NP-hard \cite{chickering1996learning}, most of the methods focus on the approximation of it. We assume we know the topological order of endogenous variables in SCM from which we get the observational data as we can get the estimated topological order by using the algorithm such as SCORE \cite{rolland2022score} or DiffAN \cite{sanchez2022diffusion}, which use the properties of the leaf nodes and iteratively extract the leaf nodes to construct the topological order from the observational data.

Furthermore, we introduce \textit{do-operator} that represents the intervention on SCM $\calM$ as follows.

\begin{definition}[do-operator]
\label{def_do_operator}
For any $i \in [d]$, We define $\doop(X_i = x_i)$ by setting the corresponding exogenous variable to the intervened value $U_i = x_i$ and deleting all the edges coming into $X_i$ from the endogenous variables on SCM.
\end{definition}
Fig. \ref{fig_do} shows the example of the do-operator where we intervene in the endogenous variable $X_2$ to $x_2$ on the SCM in Fig. \ref{fig_dag_scm}.

The following defines the \textit{average treatment effect (ATE)}, one of the causal effects we are interested in, using the do-operator. 

\begin{definition}[Average Treatment Effect]
\label{def_ate}
For all $X_i, X_j \in \calV$ in SCM such that $i \neq j$, we define the ATE of the variable (cause) $X_i$ on the variable (outcome) $X_j$ when we compare two counterfactual situations $X_i = x_i$ and $X_i = 0$ by 
\begin{align*}
    ATE(x_i, 0)
    & := \mE[x_j \given do(X_i = x_i)] - \mE[x_j \given do(X_i = 0)] \\
    & = \int_{x_j} x_j \nu(X_j = x_j \given \doop(X_i = x_i)) dx_j \\
    & - \int_{x_j} x_j \nu(X_j = x_j \given \doop(X_i = 0)) dx_j
\end{align*}
where $\nu(X_j \given \doop(X_i = x_i))$ is the probability density function of $X_j$ after the surgery on the SCM by do operator $\doop(X_i = x_i)$.
\end{definition}

\begin{figure}[t]
    \includegraphics[scale=0.25]{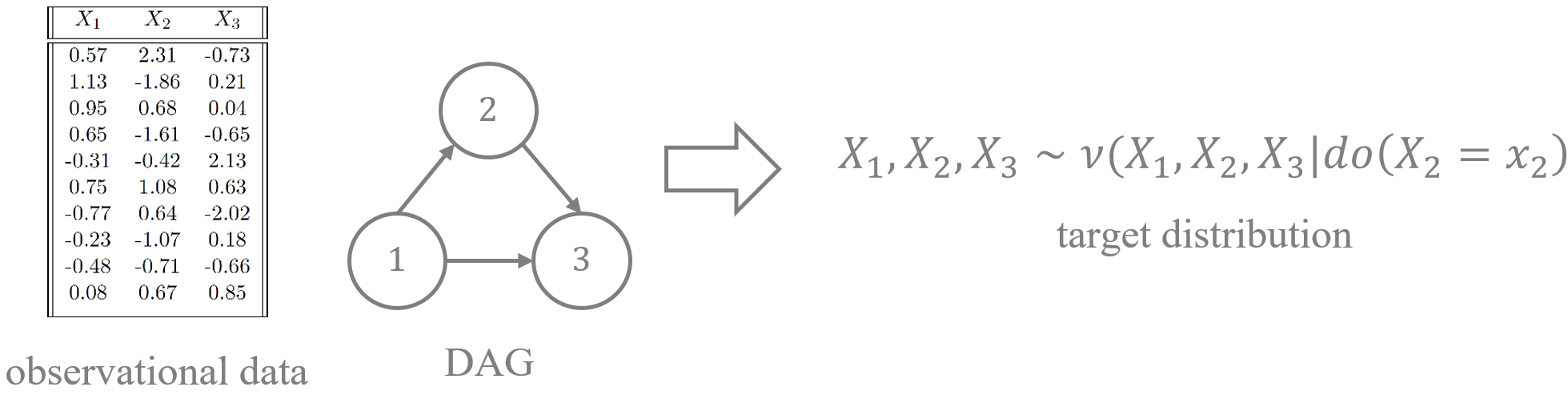}
    \centering
    \caption{Illustration of our problem: sampling from the target distribution after the intervention on a node in the SCM given the observational data and the underlying DAG}
    \label{fig_problem}
\end{figure}

As we aim to figure out the causal effect of an arbitrary node on an arbitrary node, our problem boils down to how to approximately sample from the target distribution $\nu(X_j \given \doop(X_i = x_i))$ given observational data $X \in \mR^{n \times d}$ and underlying DAG for all $i, j \in [d]$ such that $i \neq j$ shown in Fig. \ref{fig_problem}. Note that we can estimate the underlying DAG by the topological order $\pi$ and edge pruning algorithm \cite{buhlmann2014cam} that uses the feature selection.

\section{Existing Algorithm}
We introduce a diffusion-based algorithm called Diffusion-based Causal Model (DCM) proposed by Chao et al. (2023) \cite{chao2023interventional}, that can sample from the target distribution $\nu(X_j \given \doop(X_i = x_i))$ more accurately than existing state-of-the-art algorithms \cite{sanchez2021vaca} and \cite{khemakhem2021causal} under the following causal sufficiency. 

\begin{assumption}[Causal Sufficiency]
We say that the data-generating process satisfies causal sufficiency if no unmeasured confounders exist.
\label{ass_causal_sufficiency}
\end{assumption}

DCM uses Denoising Diffusion Implicit Model (DDIM) \cite{song2020denoising}, a more efficient sampling algorithm than Denoising Diffusion Probabilistic Model (DDPM) \cite{sohl2015deep} \cite{ho2020denoising}, which attained the groundbreaking performance in generating image and audio data \cite{kong2020diffwave}, \cite{ramesh2022hierarchical}, \cite{saharia2022photorealistic}. DCM trains the diffusion model at each node to capture the characteristics of the exogenous nodes in SCM. In the forward diffusion process for each endogenous node, where we gradually add the isotropic Gaussian noise, we obtain the standard Gaussian distribution. Then, in the reverse diffusion process, we decode it by adding the Gaussian distribution with a learned parameter $\theta$ to sample from the target distribution. As \cite{luo2022understanding} shows that learning the parameter in the reverse diffusion process is equivalent to learning how much noise we add at each step, we also construct the neural network that captures how much noise $\epsilon$ we should add according to the time $t$ and the already sampled values of the parent nodes $\hat{X}_{\pa_i}$ where $X_{\pa_i}$ is the set of the parent nodes of $X_i$ in SCM $\calM$. After the training, we can sample from the target distribution. We sample the root node $X_i$ in SCM from the empirical distribution $E_i$. For the intervened node $X_i$, we set it to the intervened value $\gamma_i$. For other nodes $X_i$, we sample by the reverse diffusion process $\Dec_i(Z_i, \pa(X_i))$ using the trained neural network $\epsilon_\theta$ with parent nodes $\pa(X_i)$ and the corresponding proy exogenous nodes $Z_i \sim \calN(0, 1)$. Algorithms \ref{algo_dec}, \ref{algo_DCM_train}, and \ref{algo_DCM_sample} show the comprehensive procedure of decoding, training, and sampling processes, respectively.

\begin{algorithm}[t]
\caption{$\Dec_i(Z_i, X_{\pa_i})$ \cite{chao2023interventional}}
    \begin{algorithmic}
    \STATE Input: $Z_i, X_{\pa_i}$
    \STATE Sample $\hat{X}^T \sim Z_i$
    \FOR{$t = T, \cdots, 1$}
    \STATE 
    \begin{align*}
        \hat{X}_i^{t-1} \gets \sqrt{\frac{\alpha_{t-1}}{\alpha_t}} \hat{X}_i^t - \epsilon_{\theta}^i(\hat{X}_i^t, X_{\pa_i}, t) \\
        \times \left( \sqrt{\frac{\alpha_{t-1}(1-\alpha_t)}{\alpha_t}} - \sqrt{1 - \alpha_{t-1}} \right)
    \end{align*}
    \ENDFOR
    \STATE Output: $\hat{X}_i^0$
    \end{algorithmic}
    \label{algo_dec}
\end{algorithm}

\begin{algorithm}[t]
\caption{DCM Training \cite{chao2023interventional}}
    \begin{algorithmic}
    \STATE Input: target distribution $\nu$, scale factors $\{\alpha_t\}_{t=1}^T$, DAG $\calG$ whose node $i$ is represented by $X_i$
    \WHILE{not converge}
    \STATE Sample $X^0 \sim \nu$
    \FOR{$i = 1, \cdots, d$}
    \STATE $t \sim \text{Unif}[\{1, \cdots, T\}]$
    \STATE $\epsilon \sim \calN(0, 1)$
    \STATE Update the parameter of the node $i$'s diffusion model $\epsilon_{\theta}^i$ by minimization of the following loss function by Adam optimizer 
    \begin{align*}
        \left\| \epsilon - \epsilon_{\theta}^i \left( \sqrt{\alpha_t} X_i^0 + \sqrt{1 - \alpha_t} \epsilon, X^0_{\pa_i}, t \right) \right\|_2^2
    \end{align*}
    \ENDFOR
    \ENDWHILE
    \end{algorithmic}
    \label{algo_DCM_train}
\end{algorithm}

\begin{algorithm}
\caption{DCM Sampling \cite{chao2023interventional}}[t]
    \begin{algorithmic}
    \STATE Input: Intervened node $j$ with value $\gamma_j$, noise $Z_i \sim \calN(0, 1)$ for all $i \in [d]$
    \FOR{$i = 1, \cdots, d$}
    \IF{$i$ is a root node}
    \STATE $\hat{X}_i \sim E_i$
    \ELSIF{$i = j$}
    \STATE $\hat{X}_i \gets \gamma_i$
    \ELSE
    \STATE $\hat{X}_i \gets \Dec_i\left(Z_i, \hat{X}_{\pa_i}\right)$
    \ENDIF
    \ENDFOR
    \RETURN $\hat{X} = \left(\hat{X}_1, \cdots, \hat{X}_d\right)$
    \end{algorithmic}
    \label{algo_DCM_sample}
\end{algorithm}
One of the crucial limitations of DCM \cite{chao2023interventional} is that we cannot cope with the situation where there exist unmeasured confounders, which is often the case with the data collection for business, public health and social science where causal inference makes a significant contribution. 

\section{Proposed Algorithm}
\subsection{Backdoor Diffusion-based Causal Model}
To overcome the problem of DCM and use the observational data as much as possible, we introduce the novel \textit{\textbf{B}ackdoor Criterion-based \textbf{DCM} (\textbf{BDCM})} algorithm inspired by the backdoor criterion proposed by Pearl \cite{pearl2016causal}. To define the backdoor criterion, we introduce the notion of blocking a path in DAG.

\begin{definition}[Block a Path]
    \label{def_block_a_path}
    We say that the node $Z$ blocks a path $P$ if the path $P$ includes a \textit{chain} $L \to Z \to R$, or a \textit{folk} $L \gets Z \to R$ where $L$ and $R$ are the nodes in the path $P$.
\end{definition}

 Then, using Definition \ref{def_block_a_path}, we define \textit{backdoor criterion} as follows.

\begin{definition}[Backdoor Criterion]
\label{backdoor}
A set of variables $\calB$ satisfies backdoor criterion \cite{pearl2016causal} for tuple $(X, Y)$ in DAG $\calG$ if no node in $\calB$ is a descendant of $X$ and 
$\calB$ blocks all paths between $X$ (cause) and $Y$ (outcome) which contains an arrow into $X$.
\end{definition}

\begin{algorithm}[t]
\caption{BDCM Training}
    \begin{algorithmic}
    \STATE Input: target distribution $\nu$, scale factors $\{\alpha_t\}_{t=1}^T$, DAG $\calG$ whose node $i$ is represented by $X_i$ and intervened node $j$ with intervened value $\gamma_j$
    \WHILE{not converge}
    \STATE Sample $X^0 \sim \nu$
    \FOR{$i = 1, \cdots, d$}
    \STATE $t \sim \text{Unif}[\{1, \cdots, T\}]$
    \STATE $\epsilon \sim \calN(0, 1)$
    \STATE Update the parameter of the node $i$'s diffusion model $\epsilon_{\theta}^i$ by minimization of the following loss function depending on the nodes.
    \IF{$X_j \in X_{Pa_i}$}
    \STATE
    \begin{align*}
        \left\| \epsilon - \epsilon_{\theta}^i \left( \sqrt{\alpha_t} X_i^0 + \sqrt{1 - \alpha_t} \epsilon, X_{\calB_i}^0, X_j, t \right) \right\|_2^2 \quad 
    \end{align*}
    \ELSE
    \STATE
    \begin{align*}
        \left\| \epsilon - \epsilon_{\theta}^i \left( \sqrt{\alpha_t} X_i^0 + \sqrt{1 - \alpha_t} \epsilon, X_{\calB_i}^0, t \right) \right\|_2^2 \quad 
    \end{align*}
    \ENDIF
    \ENDFOR
    \ENDWHILE
    \end{algorithmic}
    \label{algo_BDCM_train}
\end{algorithm}

If unmeasured confounders exist, then the Backdoor criterion tells us which variable to adjust concerning tuple $(X, Y)$. Then, the idea of Backdoor DCM is that for each node $X_i$ in SCM, instead of having the parents $X_{Pa_i}$ and corresponding exogenous nodes $Z_i$ as the input of the decoder of the diffusion model, we include the nodes which meet the backdoor criterion $X_{\calB_i}$ and the corresponding exogenous nodes $Z_i$ and also include the intervened node $X_j$ if it is the child of the intervened node ($X_j \in X_{Pa_i}$). Furthermore, we change the training process accordingly. As the parent nodes of the outcome node always satisfy the backdoor criterion under Assumption \ref{ass_causal_sufficiency} \cite{pearl2016causal}, including the nodes that meet the backdoor criterion instead of the parent nodes in the decoder of BDCM is the generalized algorithm of DCM. Algorithms \ref{algo_BDCM_train} and \ref{algo_BDCM_sample} show the training and sampling process of BDCM. Then, we have the following conjecture.

\begin{algorithm}[t]
\caption{BDCM Sampling}
    \begin{algorithmic}
    \STATE Input: Intervened node $j$ with value $\gamma_j$, noise $Z_i \sim \calN(0, 1)$ for all $i \in [d]$
    \FOR{$i = 1, \cdots, d$}
    \IF{$i = j$}
    \STATE $\hat{X}_i \gets\gamma_i$
    \ELSIF{$i$ is a root node}
    \STATE $\hat{X}_i \sim E_i$
    \ELSIF{$X_j \in X_{Pa_i}$}
    \STATE $\hat{X}_i \gets \Dec_i\left(Z_i, \hat{X}_{\calB_i}, X_j\right)$
    \ELSE
    \STATE $\hat{X}_i \gets \Dec_i\left(Z_i, \hat{X}_{\calB_i}\right)$
    \ENDIF
    \ENDFOR
    \RETURN $\hat{X} = \left(\hat{X}_1, \cdots, \hat{X}_d\right)$
    \end{algorithmic}
    \label{algo_BDCM_sample}
\end{algorithm}

\begin{conjecture}[Applicability of BDCM]
Suppose sets of nodes satisfy the backdoor criterion for the intervened node and other nodes. In that case, we can generalize DCM to BDCM to sample from the target distribution even if Assumption \ref{ass_causal_sufficiency} is violated.
\label{conj1}
\end{conjecture}

\subsection{Experiment}

To show that BDCM precisely samples from the target distribution where we cannot use DCM, we conduct an empirical analysis with the following settings where causal sufficiency does not hold. Python code for the experiment is available in \href{https://github.com/tatsu432/BDCM}{https://github.com/tatsu432/BDCM}.

\begin{figure}[t]
    \includegraphics[scale=0.3]{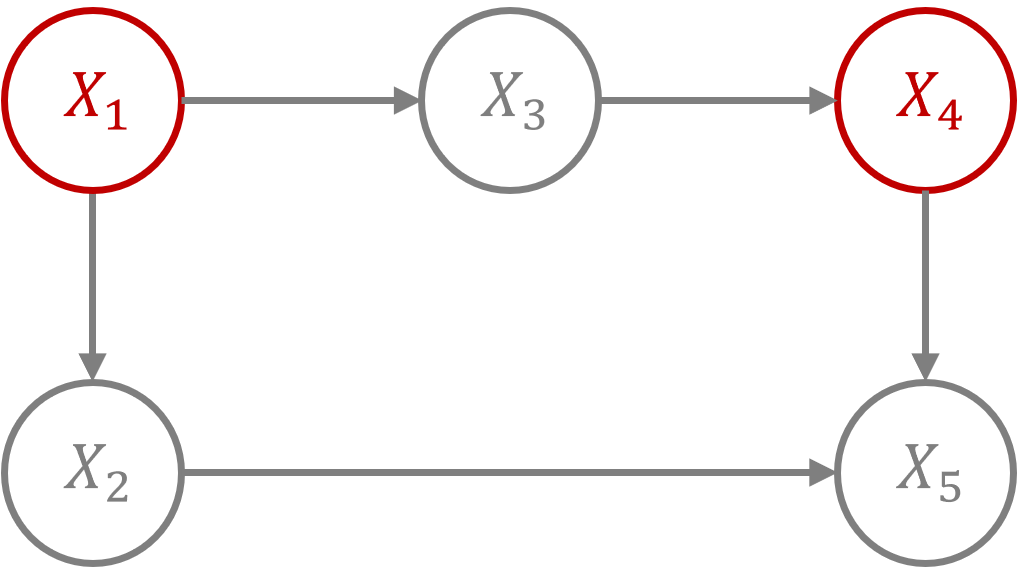}
    \centering
    \caption{SCM $\calM_1$ where the unobserved confounders $X_1$ and $X_4$ exist with five exogenous and endogenous nodes where we intervene in the node $X_2 = x_2$}

    \label{fig_experiment_1st_SCM}
\end{figure}

\begin{figure}[t]
    \includegraphics[scale=0.38]{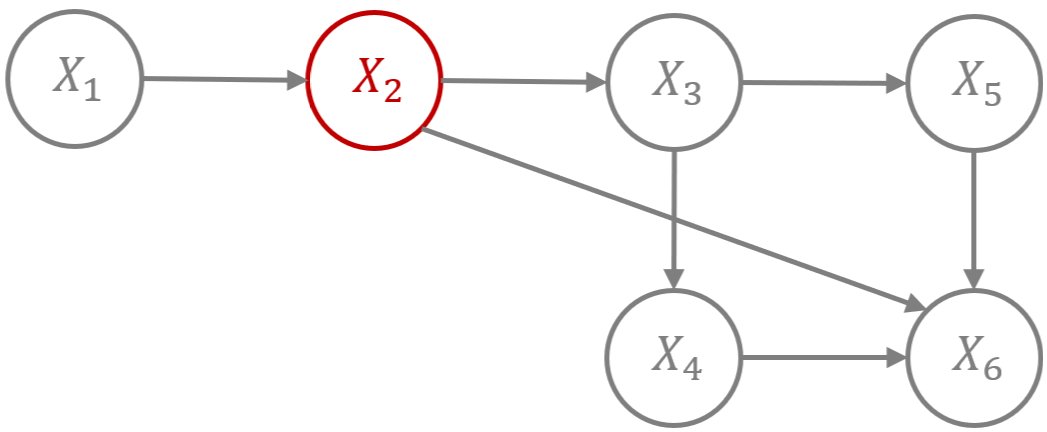}

    \caption{SCM $\calM_2$ where the unobserved confounder $X_2$ exists with six exogenous and endogenous nodes where we intervene in the node $X_4 = x_4$}

    \label{fig_experiment_2nd_SCM}
\end{figure}

Fig. \ref{fig_experiment_1st_SCM} and Fig. \ref{fig_experiment_2nd_SCM} show the SCMs $\calM_1$ and $\calM_2$ respectively that do not satisfy Assumption \ref{ass_causal_sufficiency} where $X_1$ and $X_4$ in Fig. \ref{fig_experiment_1st_SCM} and $X_2$ in Fig. \ref{fig_experiment_2nd_SCM} are the unobserved nodes. Note that we did not show the exogenous nodes in the figures for clarity. Examples \ref{ex_SCM1_simple} and \ref{ex_SCM1_complex} show the concrete structural equations for $\calM_1$ and Examples \ref{ex_SCM2_simple} and \ref{ex_SCM2_complex} for $\calM_2$. We create simple and complex structural equations for both cases. The simple cases are the \textit{additive noise models (ANM)} \cite{shimizu2006linear}, \cite{hoyer2008nonlinear}, \cite{peters2014causal}, \cite{buhlmann2014cam} whereas the complex ones are not ANM.

\begin{example}
We define the set of simple structural equations $f=\{f_i\}_{i \in [5]}$ for SCM $\calM_1$ in Fig. \ref{fig_experiment_1st_SCM} as follows.
\begin{align*}
  X_1 &= f_1(U_1) = U_1 \\
  X_2 &= f_2(X_1, U_2) = X_1^2 + U_2 \\
  X_3 & = f_3(X_1, U_3) = 2 X_1 + U_3 \\
  X_4 & = f_4(X_3, U_4) = X_3 + U_4 \\
  X_5 &= f_5(X_2, X_4, U_5) = X_2 + 2 X_4 + U_5
\end{align*}
\label{ex_SCM1_simple}
\end{example}

\begin{example}
We define the set of complex structural equations $f=\{f_i\}_{i \in [5]}$ for SCM $\calM_1$ in Fig. \ref{fig_experiment_1st_SCM} as follows.
\begin{align*}
      X_1 &= f_1(U_1) = U_1 \\
      X_2 &= f_2(X_1, U_2) = \frac{\sqrt{|X_1|} (|U_2| + 0.1) }{2}+ |X_1| + \frac{U_2}{5} \\
      X_3 & = f_3(X_1, U_3) = \frac{1}{1 + (|U_3| + 0.1) \exp(-X_2)} \\
      X_4 & = f_4(X_3, U_4) = X_3 + X_3 U_4 + U_4 \\
      X_5 &= f_5(X_2, X_4, U_5) = X_2 + X_4 + X_2 X_4 U_5 + U_5
\end{align*}
\label{ex_SCM1_complex}
\end{example}

\begin{example}
We define the set of simple structural equations $f=\{f_i\}_{i \in [6]}$ for SCM $\calM_2$ in Fig. \ref{fig_experiment_2nd_SCM} as follows.
\begin{align*}
  X_1 &= f_1(U_1) = U_1 \\
  X_2 &= f_2(X_1, U_2) = X_1^2 + U_2 \\
  X_3 & = f_3(X_2, U_3) = X_2 + U_3 \\
  X_4 & = f_4(X_3, U_4) = X_3^3 + X_3 + U_4 \\
  X_5 &= f_5(X_3, U_5) = X_3^2 + 0.1 + U_5 \\
  X_6 &= f_6(X_2, X_4, X_5, U_6) = X_2 X_4 + X_2 X_5 + X_4 X_5 + U_6
\end{align*}
\label{ex_SCM2_simple}
\end{example}

\begin{example}
We define the set of complex structural equations $f=\{f_i\}_{i \in [6]}$ for SCM $\calM_2$ in Fig. \ref{fig_experiment_2nd_SCM} as follows.
\begin{align*}
  X_1 &= f_1(U_1) = U_1 \\
  X_2 &= f_2(X_1, U_2) = \frac{\sqrt{|X_1|} (|U_2| + 0.1)}{2} + |X_1| + \frac{U_2}{5} \\
  X_3 & = f_3(X_2, U_3) = \frac{1}{1 + (|U_3| + 0.1) \exp(-X_2)} \\
  X_4 & = f_4(X_3, U_4) = \frac{U_4 (|X_3| + 0.3)}{5} + U_4 \\
  X_5 &= f_5(X_3, U_5) = \frac{1}{\sqrt{|U_5 X_3|} + 0.1} + U_5 \\
  X_6 &= f_6(X_2, X_4, X_5, U_6) \\
  &= X_2^2 X_4 + X_2 X_5 + X_5 X_6 + X_2 U_6
\end{align*}
\label{ex_SCM2_complex}
\end{example}

For Examples \ref{ex_SCM1_simple}, \ref{ex_SCM1_complex}, \ref{ex_SCM2_simple}, and \ref{ex_SCM2_complex}, we sample exogenous nodes $U_i$ from standard normal distribution $\calN(0, 1)$ for all $i \in [5]$ in $\calM_1$ and all $i \in [6]$ in $\calM_2$. We normalized each endogenous variable as \cite{chao2023interventional} did. 

For both Examples \ref{ex_SCM1_simple} and \ref{ex_SCM1_complex} for Fig. \ref{fig_experiment_1st_SCM}, we aim to sample correctly from the target distribution $\nu(X_5|\doop(X_2 = x_2))$ where $X_2$ is the cause, and $X_5$ is the outcome. For both DCM and BDCM, we set the intervened node $X_2$ to intervened value $x_2$ and sample $X_3$ from the empirical distribution $E_3$. For the node of our interest $X_5$, DCM takes $\hat{X}_2$ as the input for the decoder $\Dec_5(Z_5, \hat{X}_2)$ whereas BDCM takes $\hat{X}_2$ and $\hat{X}_3$ as the input for the decoder $\Dec_5(Z_5, \hat{X}_2, \hat{X}_3)$.

For both Examples \ref{ex_SCM2_simple} and \ref{ex_SCM2_complex} for Fig. \ref{fig_experiment_2nd_SCM}, we aim to sample correctly from the target distribution $\nu(X_6|\doop(X_4 = x_4))$ where $X_4$ is the cause, and $X_6$ is the outcome. For both DCM and BDCM, we set the intervened node $X_4$ to intervened value $x_4$, sample $X_1$ and $X_3$ from the empirical distribution $E_1$ and $E_3$ respectively, and sample $X_5$ by the decoder $\Dec_5(Z_5, \hat{X}_3)$. For the node of our interest $X_6$, DCM takes $\hat{X}_4$ and $\hat{X}_5$ as the inputs for the decoder $\Dec_6(Z_6, \hat{X}_4, \hat{X}_5)$ whereas BDCM takes $\hat{X}_3$ and $\hat{X}_4$ as the inputs for the decoder $\Dec_6(Z_6, \hat{X}_3, \hat{X}_4)$.

For parameters in the algorithm, we set them to the following values, mostly the same as \cite{chao2023interventional}. For the noise schedule $\beta_t$ and $\alpha_t$, we set them to $\beta_t = \left(0.1 - 10^{-4}\right) \frac{t - 1}{T - 1} + 10^{-4}$ and $\alpha_t = \prod_{i = 1}^t (1 - \beta_t)$ where we set $T = 100$. For the neural networks, we set the epochs to $500$, batch size to $64$, and learning rate to $10^{-4}$ where each neural network consists of three hidden layers whose numbers of nodes are $128, 256$, and $256$ for the first, second and third layers, respectively. We extract $500$ samples via DCM and BDCM, where we train them with $1000$ samples. We calculate the Maximum Mean Discrepancy (MMD) \cite{gretton2012kernel} between the empirical distributions obtained from the algorithms and the ground truth target for both DCM and BDCM. Note that the lower MMD is, the closer the empirical distributions are, so the algorithm is more precise. We set the intervened values to ten different values sampled randomly from $\text{Unif}(-3, 3)$. We also conduct the simulation for five different seeds. Then, We output the average and standard deviation of MMDs.

\begin{table}[t]

    \caption{Average $\pm$ standard deviation of MMD $(\times 10^{-3})$ of DCM and BDCM compared to the true target distribution in Examples \ref{ex_SCM1_simple}, \ref{ex_SCM1_complex}, \ref{ex_SCM2_simple}, \ref{ex_SCM2_complex}, \ref{ex_SCM3_simple}, \ref{ex_SCM3_complex}, \ref{ex_SCM4_simple}, \ref{ex_SCM4_complex}, \ref{ex_SCM5_simple}, \ref{ex_SCM5_complex}}
    \begin{tabular}[t]{cccc}
        \toprule
                      &&BDCM (ours)&DCM\\
        \midrule
        \multirow{2}{*}{SCM $\calM_1$} 
        &Ex. \ref{ex_SCM1_simple}&$\mathbf{1.24 \pm 0.744}$&$1.79 \pm 1.54$\\
        &Ex. \ref{ex_SCM1_complex}&$\mathbf{1.04 \pm 0.835}$&$2.34 \pm 2.17$\\
        \midrule
        \midrule
        
        \multirow{2}{*}{SCM $\calM_2$} 
        &Ex. \ref{ex_SCM2_simple}&$5.08 \pm 2.51$&$5.07 \pm 2.17$\\
        &Ex. \ref{ex_SCM2_complex}&$\mathbf{1.55 \pm 1.91}$&$2.89 \pm 2.08$\\
        \midrule
        \midrule
        
        \multirow{2}{*}{SCM $\calM_3$} 
        &Ex.\ref{ex_SCM3_simple}&$\mathbf{0.741 \pm 0.68}$&$1.14\pm 1.26$\\
        &Ex.\ref{ex_SCM3_complex}&$\mathbf{1.51 \pm 1.43}$&$1.8 \pm 1.55$\\
        \midrule
        \midrule
        
        \multirow{2}{*}{SCM $\calM_4$} 
        &Ex. \ref{ex_SCM4_simple}&$\mathbf{1.69 \pm 1.49}$&$2.12 \pm 1.34$\\
        &Ex. \ref{ex_SCM4_complex}&$\mathbf{1.46 \pm 1.11}$&$2.38 \pm 1.81$\\
        \midrule
        \midrule
        
        \multirow{2}{*}{SCM $\calM_5$} 
        &Ex. \ref{ex_SCM5_simple}&$\mathbf{0.638 \pm 0.586}$&$0.747 \pm 0.575$\\
        &Ex. \ref{ex_SCM5_complex}&$\mathbf{1.29 \pm 0.938}$&$1.41 \pm 0.591$\\
        \bottomrule
    \end{tabular}
    \label{table_mmd}
\end{table}
Table \ref{table_mmd} shows the results of the experiments. Table \ref{table_mmd} demonstrates that BDCM output a more precise distribution than DCM, where unmeasured confounders exist for Examples \ref{ex_SCM1_simple}, \ref{ex_SCM1_complex}, \ref{ex_SCM2_complex}. For Example \ref{ex_SCM2_simple}, BDCM is almost as accurate as DCM. For both SCMs $\calM_1$ and $\calM_2$, the more complex the structural equations become in SCM, the clear the difference in the performance between DCM and BDCM is. For SCM $\calM_1$ in Fig. \ref{fig_experiment_1st_SCM}, BDCM successfully considers the backdoor path $X_2 \gets X_1 \to X_3 \to X_4 \to X_5$ by including the node $X_3$ that blocks the backdoor path in the decoder of the outcome meanwhile DCM does not consider this path when we sample the outcome $X_5$ where we intervene in the node $X_2$, which creates the bias. Furthermore, for SCM $\calM_2$ in Fig. \ref{fig_experiment_2nd_SCM}, BDCM carefully chooses the nodes $X_3$ and $X_4$ that block all the backdoor paths concerning the pair of the cause and outcome nodes as the input for the decoder of the outcome $X_6$ of our interest. In contrast, DCM takes the parent nodes of the outcome we observe $X_4$ and $X_5$ without considering one of the backdoor paths: $X_4 \gets X_3 \gets X_2 \to X_6$, which incurs the bias in the sample by DCM. 

\begin{figure}[t]
    \includegraphics[scale=0.335]{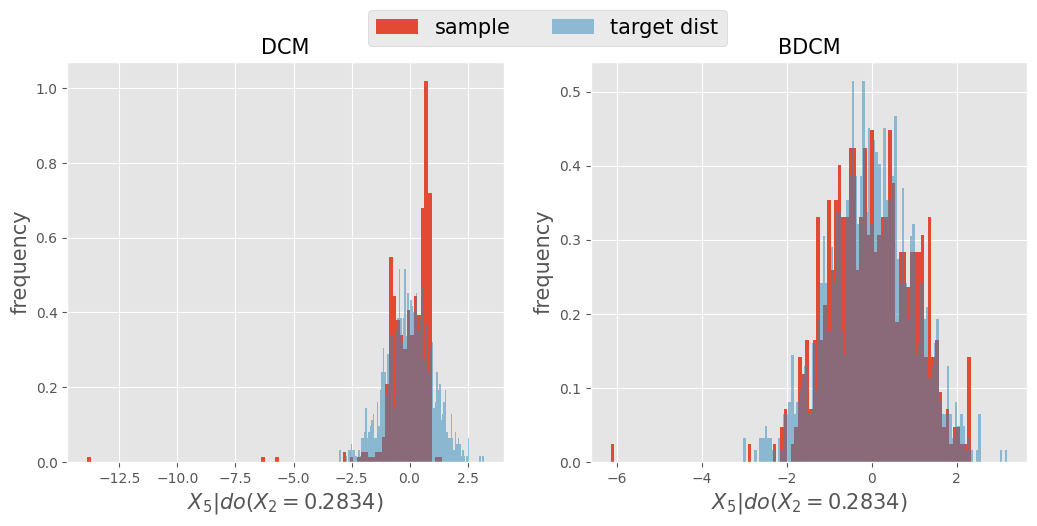}
    \centering
    \caption{Empirical distributions of the $X_5$ sampled from DCM (left) and BDCM (right) compared to the ground-truth target distribution where we intervened in the node $X_2 = 0.2834$ in Example \ref{ex_SCM1_simple}}

    \label{fig_res_SCM1_simple}
\end{figure}

\begin{figure}[t]
    \includegraphics[scale=0.335]{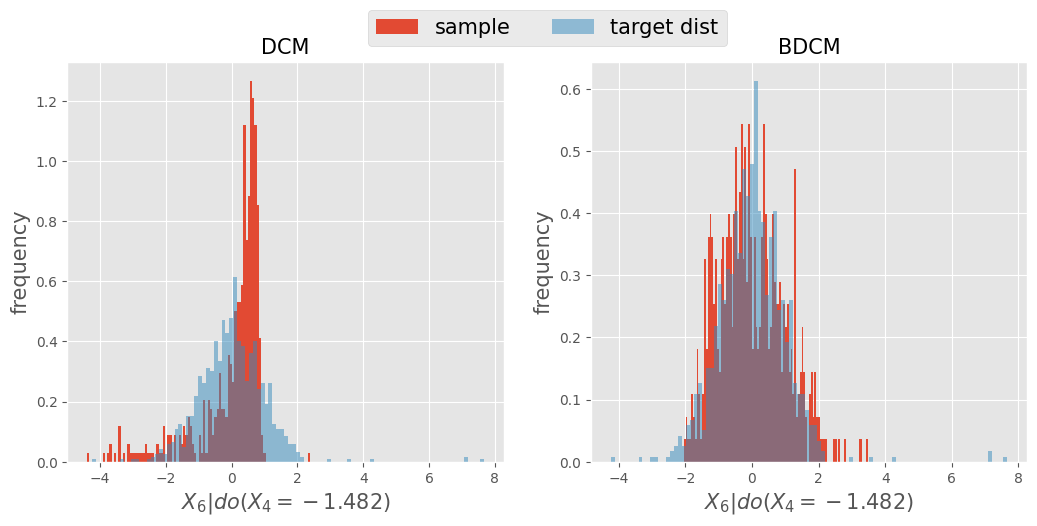}
    \centering
    \caption{Empirical distributions of the $X_6$ sampled from DCM (left) and BDCM (right) compared to the ground-truth target distribution where we intervened in the node $X_4 = -1.482$ in Example \ref{ex_SCM2_simple}}

    \label{fig_res_SCM2_simple}
\end{figure}

Furthermore, Fig. \ref{fig_res_SCM1_simple} and Fig. \ref{fig_res_SCM1_complex} show ones of the empirical distributions sampled by DCM, BDCM, and target distribution for SCMs $\calM_1$ and $\calM_2$ where the structural equations are complex. The blue histograms are the ground truth distribution we want to sample from, whereas the red histograms are the outputs of the DCM (left) and BDCM (right). From Fig. \ref{fig_res_SCM1_simple} and Fig. \ref{fig_res_SCM2_simple}, we can see that BDCM can sample from the target distribution $\nu(X_5|\doop(X_2 = x_2))$ in $\calM_1$ and $\nu(X_6|\doop(X_4 = x_4))$ in $\calM_2$ precisely where unmeasured confounders exist whereas DCM fails to do so.

\section{Conclusion and Future Work}
We extended the Diffusion-based causal Model (DCM) proposed by \cite{chao2023interventional} to the case where unmeasured confounders exist. We proposed Backdoor Criterion-based DCM (BDCM) that can consider the unobserved confounders by including the nodes that meet the backdoor criterion \cite{pearl2016causal}. Synthetic data experiment demonstrates that BDCM can precisely sample from the target distribution of our interest where DCM fails to do so.

For future work, one of the intriguing topics would be to derive the convergence guarantee of BDCM. Implementing the comprehensive algorithm of BDCM in Python would also be interesting. Moreover, it would be intriguing to generalize BDCM using the Front-door criterion \cite{pearl2016causal}, another criterion to adjust the nodes where unobserved confounders exist.

\section*{Acknowledgements} 
We thank Dr. Andre Wibisono at Yale University Department of Computer Science for his help and guidance.



\bibliographystyle{icml2023}
\bibliography{ref}

\begin{thebibliography}{23}
\providecommand{\natexlab}[1]{#1}
\providecommand{\url}[1]{\texttt{#1}}
\expandafter\ifx\csname urlstyle\endcsname\relax
  \providecommand{\doi}[1]{doi: #1}\else
  \providecommand{\doi}{doi: \begingroup \urlstyle{rm}\Url}\fi

\bibitem[B{\"u}hlmann et~al.(2014)B{\"u}hlmann, Peters, and Ernest]{buhlmann2014cam}
B{\"u}hlmann, P., Peters, J., and Ernest, J.
\newblock Cam: Causal additive models, high-dimensional order search and penalized regression.
\newblock \emph{Ann. Statist.}, 2014.

\bibitem[Chao et~al.(2023)Chao, Bl{\"o}baum, and Kasiviswanathan]{chao2023interventional}
Chao, P., Bl{\"o}baum, P., and Kasiviswanathan, S.~P.
\newblock Interventional and counterfactual inference with diffusion models.
\newblock \emph{arXiv preprint arXiv:2302.00860}, 2023.

\bibitem[Chickering(1996)]{chickering1996learning}
Chickering, D.~M.
\newblock Learning bayesian networks is np-complete.
\newblock \emph{Learning from data: Artificial intelligence and statistics V}, pp.\  121--130, 1996.

\bibitem[Glymour et~al.(2019)Glymour, Zhang, and Spirtes]{glymour2019review}
Glymour, C., Zhang, K., and Spirtes, P.
\newblock Review of causal discovery methods based on graphical models.
\newblock \emph{Frontiers in Genetics}, 10, 2019.
\newblock ISSN 1664-8021.
\newblock \doi{10.3389/fgene.2019.00524}.

\bibitem[Gretton et~al.(2012)Gretton, Borgwardt, Rasch, Sch{\"o}lkopf, and Smola]{gretton2012kernel}
Gretton, A., Borgwardt, K.~M., Rasch, M.~J., Sch{\"o}lkopf, B., and Smola, A.
\newblock A kernel two-sample test.
\newblock \emph{The Journal of Machine Learning Research}, 13\penalty0 (1):\penalty0 723--773, 2012.

\bibitem[Ho et~al.(2020)Ho, Jain, and Abbeel]{ho2020denoising}
Ho, J., Jain, A., and Abbeel, P.
\newblock Denoising diffusion probabilistic models.
\newblock \emph{Advances in neural information processing systems}, 33:\penalty0 6840--6851, 2020.

\bibitem[Hoyer et~al.(2008)Hoyer, Janzing, Mooij, Peters, and Sch{\"o}lkopf]{hoyer2008nonlinear}
Hoyer, P., Janzing, D., Mooij, J.~M., Peters, J., and Sch{\"o}lkopf, B.
\newblock Nonlinear causal discovery with additive noise models.
\newblock \emph{Advances in neural information processing systems}, 21, 2008.

\bibitem[Imbens \& Rubin(2015)Imbens and Rubin]{imbens2015causal}
Imbens, G.~W. and Rubin, D.~B.
\newblock \emph{Causal inference in statistics, social, and biomedical sciences}.
\newblock Cambridge University Press, 2015.

\bibitem[Khemakhem et~al.(2021)Khemakhem, Monti, Leech, and Hyvarinen]{khemakhem2021causal}
Khemakhem, I., Monti, R., Leech, R., and Hyvarinen, A.
\newblock Causal autoregressive flows.
\newblock In \emph{International conference on artificial intelligence and statistics}, pp.\  3520--3528. PMLR, 2021.

\bibitem[Kong et~al.(2020)Kong, Ping, Huang, Zhao, and Catanzaro]{kong2020diffwave}
Kong, Z., Ping, W., Huang, J., Zhao, K., and Catanzaro, B.
\newblock Diffwave: A versatile diffusion model for audio synthesis.
\newblock \emph{arXiv preprint arXiv:2009.09761}, 2020.

\bibitem[Luo(2022)]{luo2022understanding}
Luo, C.
\newblock Understanding diffusion models: A unified perspective.
\newblock \emph{arXiv preprint arXiv:2208.11970}, 2022.

\bibitem[Pearl et~al.(2016)Pearl, Glymour, and Jewell]{pearl2016causal}
Pearl, J., Glymour, M., and Jewell, N.~P.
\newblock \emph{Causal inference in statistics: A primer}.
\newblock John Wiley \& Sons, 2016.

\bibitem[Peters et~al.(2014)Peters, Mooij, Janzing, and Sch{\"o}lkopf]{peters2014causal}
Peters, J., Mooij, J.~M., Janzing, D., and Sch{\"o}lkopf, B.
\newblock Causal discovery with continuous additive noise models.
\newblock \emph{Journal of Machine Learning Research}, 2014.

\bibitem[Ramesh et~al.(2022)Ramesh, Dhariwal, Nichol, Chu, and Chen]{ramesh2022hierarchical}
Ramesh, A., Dhariwal, P., Nichol, A., Chu, C., and Chen, M.
\newblock Hierarchical text-conditional image generation with clip latents.
\newblock \emph{arXiv preprint arXiv:2204.06125}, 1\penalty0 (2):\penalty0 3, 2022.

\bibitem[Rolland et~al.(2022)Rolland, Cevher, Kleindessner, Russell, Janzing, Sch{\"o}lkopf, and Locatello]{rolland2022score}
Rolland, P., Cevher, V., Kleindessner, M., Russell, C., Janzing, D., Sch{\"o}lkopf, B., and Locatello, F.
\newblock Score matching enables causal discovery of nonlinear additive noise models.
\newblock In \emph{International Conference on Machine Learning}, pp.\  18741--18753. PMLR, 2022.

\bibitem[Saharia et~al.(2022)Saharia, Chan, Saxena, Li, Whang, Denton, Ghasemipour, Gontijo~Lopes, Karagol~Ayan, Salimans, et~al.]{saharia2022photorealistic}
Saharia, C., Chan, W., Saxena, S., Li, L., Whang, J., Denton, E.~L., Ghasemipour, K., Gontijo~Lopes, R., Karagol~Ayan, B., Salimans, T., et~al.
\newblock Photorealistic text-to-image diffusion models with deep language understanding.
\newblock \emph{Advances in Neural Information Processing Systems}, 35:\penalty0 36479--36494, 2022.

\bibitem[Sanchez et~al.(2022{\natexlab{a}})Sanchez, Liu, O'Neil, and Tsaftaris]{sanchez2022diffusion}
Sanchez, P., Liu, X., O'Neil, A.~Q., and Tsaftaris, S.~A.
\newblock Diffusion models for causal discovery via topological ordering.
\newblock \emph{arXiv preprint arXiv:2210.06201}, 2022{\natexlab{a}}.

\bibitem[Sanchez et~al.(2022{\natexlab{b}})Sanchez, Voisey, Xia, Watson, O’Neil, and Tsaftaris]{sanchez2022causal}
Sanchez, P., Voisey, J.~P., Xia, T., Watson, H.~I., O’Neil, A.~Q., and Tsaftaris, S.~A.
\newblock Causal machine learning for healthcare and precision medicine.
\newblock \emph{Royal Society Open Science}, 9\penalty0 (8):\penalty0 220638, 2022{\natexlab{b}}.

\bibitem[Sanchez-Martin et~al.(2021)Sanchez-Martin, Rateike, and Valera]{sanchez2021vaca}
Sanchez-Martin, P., Rateike, M., and Valera, I.
\newblock Vaca: Design of variational graph autoencoders for interventional and counterfactual queries.
\newblock \emph{arXiv preprint arXiv:2110.14690}, 2021.

\bibitem[Shimizu et~al.(2006)Shimizu, Hoyer, Hyv{\"a}rinen, Kerminen, and Jordan]{shimizu2006linear}
Shimizu, S., Hoyer, P.~O., Hyv{\"a}rinen, A., Kerminen, A., and Jordan, M.
\newblock A linear non-gaussian acyclic model for causal discovery.
\newblock \emph{Journal of Machine Learning Research}, 7\penalty0 (10), 2006.

\bibitem[Sohl-Dickstein et~al.(2015)Sohl-Dickstein, Weiss, Maheswaranathan, and Ganguli]{sohl2015deep}
Sohl-Dickstein, J., Weiss, E., Maheswaranathan, N., and Ganguli, S.
\newblock Deep unsupervised learning using nonequilibrium thermodynamics.
\newblock In \emph{International conference on machine learning}, pp.\  2256--2265. PMLR, 2015.

\bibitem[Song et~al.(2020)Song, Meng, and Ermon]{song2020denoising}
Song, J., Meng, C., and Ermon, S.
\newblock Denoising diffusion implicit models.
\newblock \emph{arXiv preprint arXiv:2010.02502}, 2020.

\bibitem[Spirtes et~al.(2000)Spirtes, Glymour, and Scheines]{spirtes2000causation}
Spirtes, P., Glymour, C.~N., and Scheines, R.
\newblock \emph{Causation, prediction, and search}.
\newblock MIT press, 2000.

\end{thebibliography}

\newpage
\appendix
\onecolumn

\section{Details of Synthetic Data Experiment}

\begin{figure}[t]
    \includegraphics[scale=0.3]{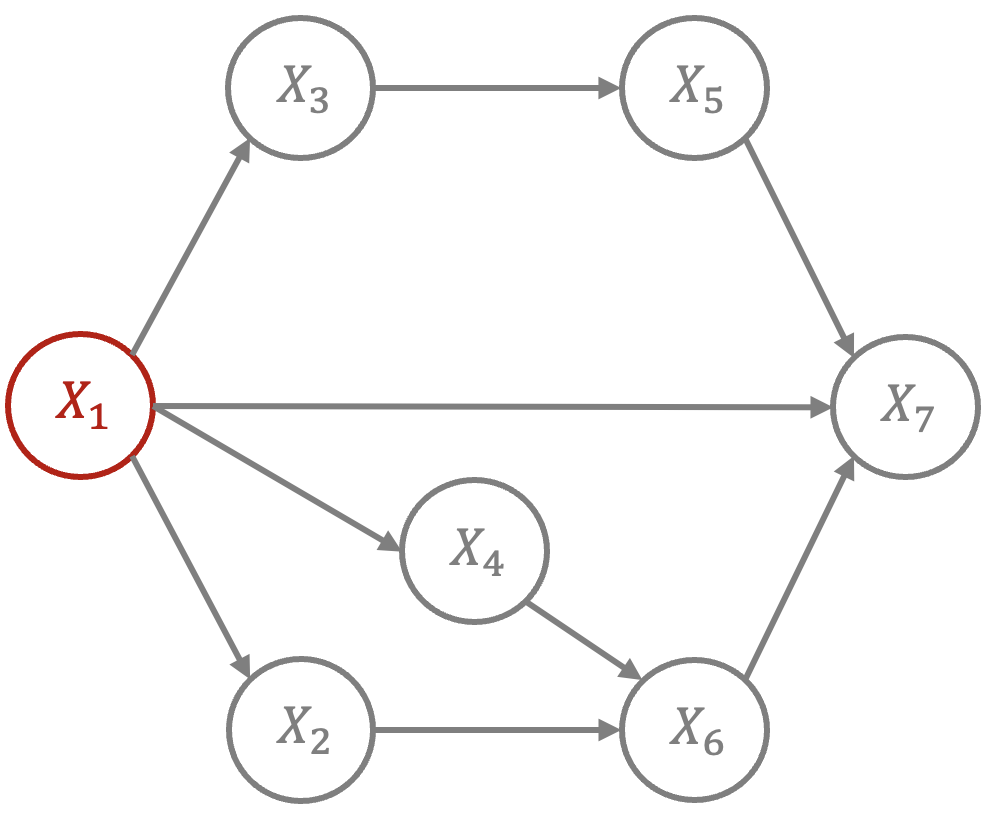}
    \centering
    \caption{SCM $\calM_3$ where the unobserved confounder $X_1$ exists with seven exogenous and endogenous nodes where we intervene in the node $X_6 = x_6$}

    \label{fig_SCM3}
\end{figure}

\begin{example}
We define the set of simple structural equations $f=\{f_i\}_{i \in [7]}$ for SCM $\calM_3$ in Fig. \ref{fig_SCM3} as follows.
\begin{align*}
  X_1 &= f_1(U_1) = U_1 \\
  X_2 &= f_2(X_1, U_2) = X_1^2 + U_2 \\
  X_3 & = f_3(X_1, U_3) = X_1 + U_3 \\
  X_4 & = f_4(X_1, U_4) = X_1^3 + X_1 + U_4 \\
  X_5 &= f_5(X_3, U_5) = X_3^2 + 0.1 + U_5 \\
  X_6 &= f_6(X_2, X_4, U_6) = X_2 X_4 + U_6 \\
  X_7 &= f_7(X_1, X_5, X_6, U_7) = X_1 X_5 + X_6^2 + X_1 X_6 + U_7 \\
\end{align*}
\label{ex_SCM3_simple}
\end{example}

\begin{example}
We define the set of complex structural equations $f=\{f_i\}_{i \in [7]}$ for SCM $\calM_3$ in Fig. \ref{fig_SCM3} as follows.
\begin{align*}
  X_1 &= f_1(U_1) = U_1 \\
  X_2 &= f_2(X_1, U_2) = \frac{\sqrt{|X_1|} (|U_2| + 0.1)}{2} + |X_1| + \frac{U_2}{5} \\
  X_3 & = f_3(X_1, U_3) = \frac{1}{1 + (|U_3| + 0.1) \exp(-X_1)} \\
  X_4 & = f_4(X_1, U_4) = \frac{U_4 (|X_3| + 0.3)}{5} + U_4 \\
  X_5 &= f_5(X_3, U_5) = \frac{1}{\sqrt{|U_5 X_3|} + 0.1} + U_5 \\
  X_6 &= f_6(X_2, X_4, U_6) \\
  &= X_2^2 X_4 + X_2 X_4 + X_2 U_6 \\
  X_7 &= f_7(X_1, X_5, X_6, U_7) = \\
  &= X_1^2 X_5 + X_1 X_6 + X_1 X_5 U_7 \\
\end{align*}
\label{ex_SCM3_complex}
\end{example}

\begin{figure}[t]
    \includegraphics[scale=0.3]{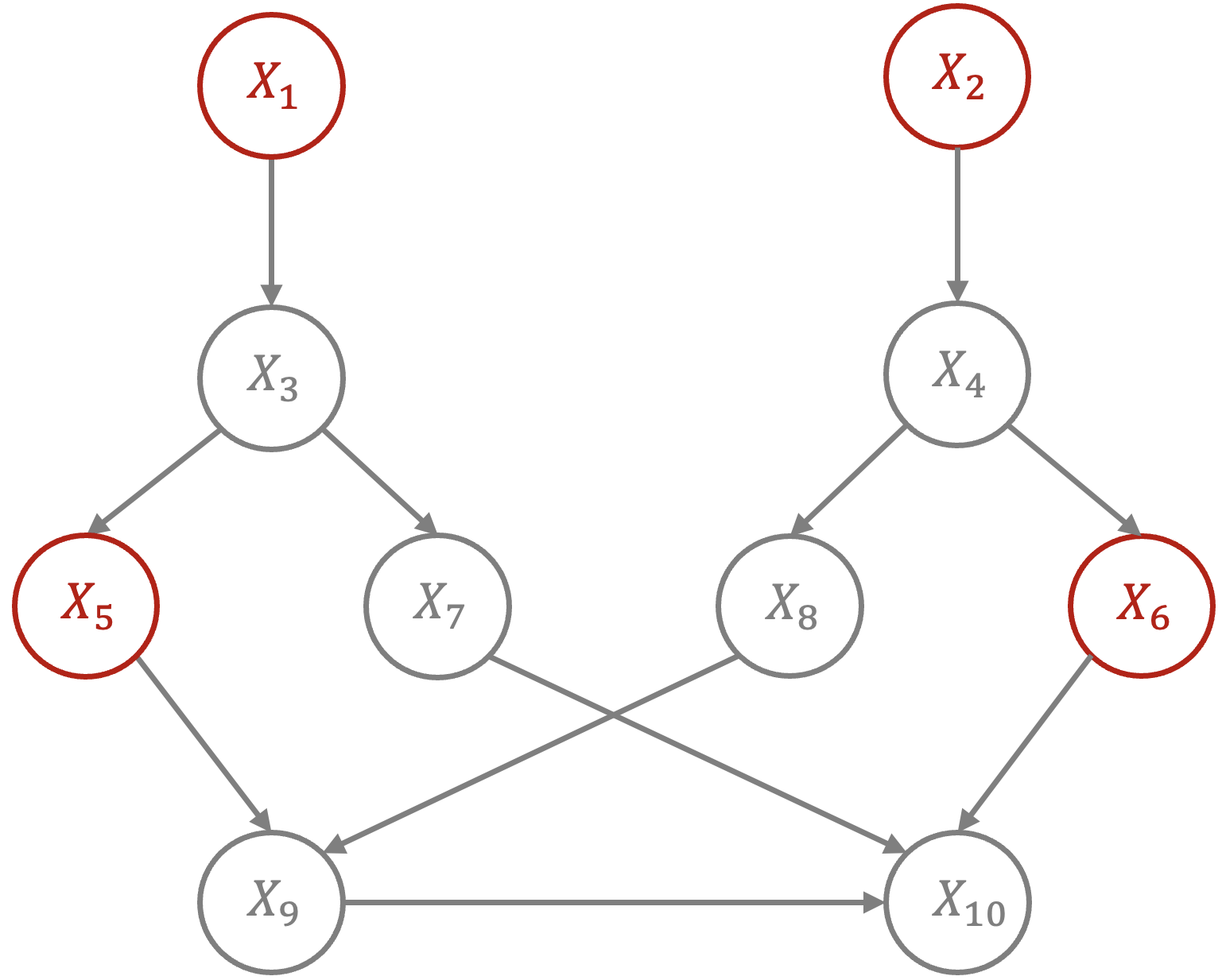}
    \centering
    \caption{SCM $\calM_4$ where the unobserved confounders $X_1$, $X_2$, $X_5$, and $X_6$ exist with five exogenous and endogenous nodes where we intervene in the node $X_9 = x_9$}

    \label{fig_SCM4}
\end{figure}

\begin{example}
We define the set of simple structural equations $f=\{f_i\}_{i \in [10]}$ for SCM $\calM_4$ in Fig. \ref{fig_SCM4} as follows.
\begin{align*}
  X_1 &= f_1(U_1) = U_1 \\
  X_2 &= f_2(U_2) = U_2 \\
  X_3 & = f_3(X_1, U_3) = X_1 + U_3 \\
  X_4 & = f_4(X_2, U_4) = -X_2^3 + X_2 + U_4 \\
  X_5 &= f_5(X_3, U_5) = X_3^2 + 0.1 + U_5 \\
  X_6 &= f_6(X_4, U_6) = X_4^2 + X_4 + U_6 \\
  X_7 &= f_7(X_3, U_7) = -X_3^2 + X_3 + U_7 \\
  X_8 &= f_8(X_4, U_8) = 3 X_4 + 0.1 + U_8 \\
  X_9 &= f_9(X_5, X_8, U_9) = X_5 X_8 + X_5 + X_8 + U_9 \\
  X_{10} &= f_{10}(X_6, X_7, X_9, U_{10}) = X_6 X_7 X_9 + X_6 X_7 + U_{10} \\
\end{align*}
\label{ex_SCM4_simple}
\end{example}

\begin{example}
We define the set of complex structural equations $f=\{f_i\}_{i \in [10]}$ for SCM $\calM_4$ in Fig. \ref{fig_SCM4} as follows.
\begin{align*}
  X_1 &= f_1(U_1) = U_1 \\
  X_2 &= f_2(U_2) = U_2 \\
  X_3 & = f_3(X_1, U_3) = \frac{\sqrt{|X_1|} (|U_3| + 0.1)}{2} + |X_1| + \frac{U_3}{5} \\
  X_4 & = f_4(X_2, U_4) = \frac{U_4 (|X_2| + 0.3)}{5} + U_4 \\
  X_5 &= f_5(X_3, U_5) = -\frac{1}{1 + (|U_5| + 0.1) \exp(-X_3)} \\
  X_6 &= f_6(X_4, U_6) = \frac{U_6 (|X_4| + 0.3)}{5} + U_6 \\
  X_7 &= f_7(X_3, U_7) = \frac{\sqrt{|X_3|} (|U_7| + 0.1)}{2} + |X_3| + \frac{U_7}{5} \\
  X_8 &= f_8(X_4, U_8) = 3 X_4 + 0.1 + U_8 \\
  X_9 &= f_9(X_5, X_8, U_9) = X_5^2 X_8 + X_5 + X_8 + U_9 \\
  X_{10} &= f_{10}(X_6, X_7, X_9, U_{10}) = X_6^2 X_7 X_9 + X_6 X_7 + U_{10} \\
\end{align*}
\label{ex_SCM4_complex}
\end{example}

\begin{figure}[t]
    \includegraphics[scale=0.3]{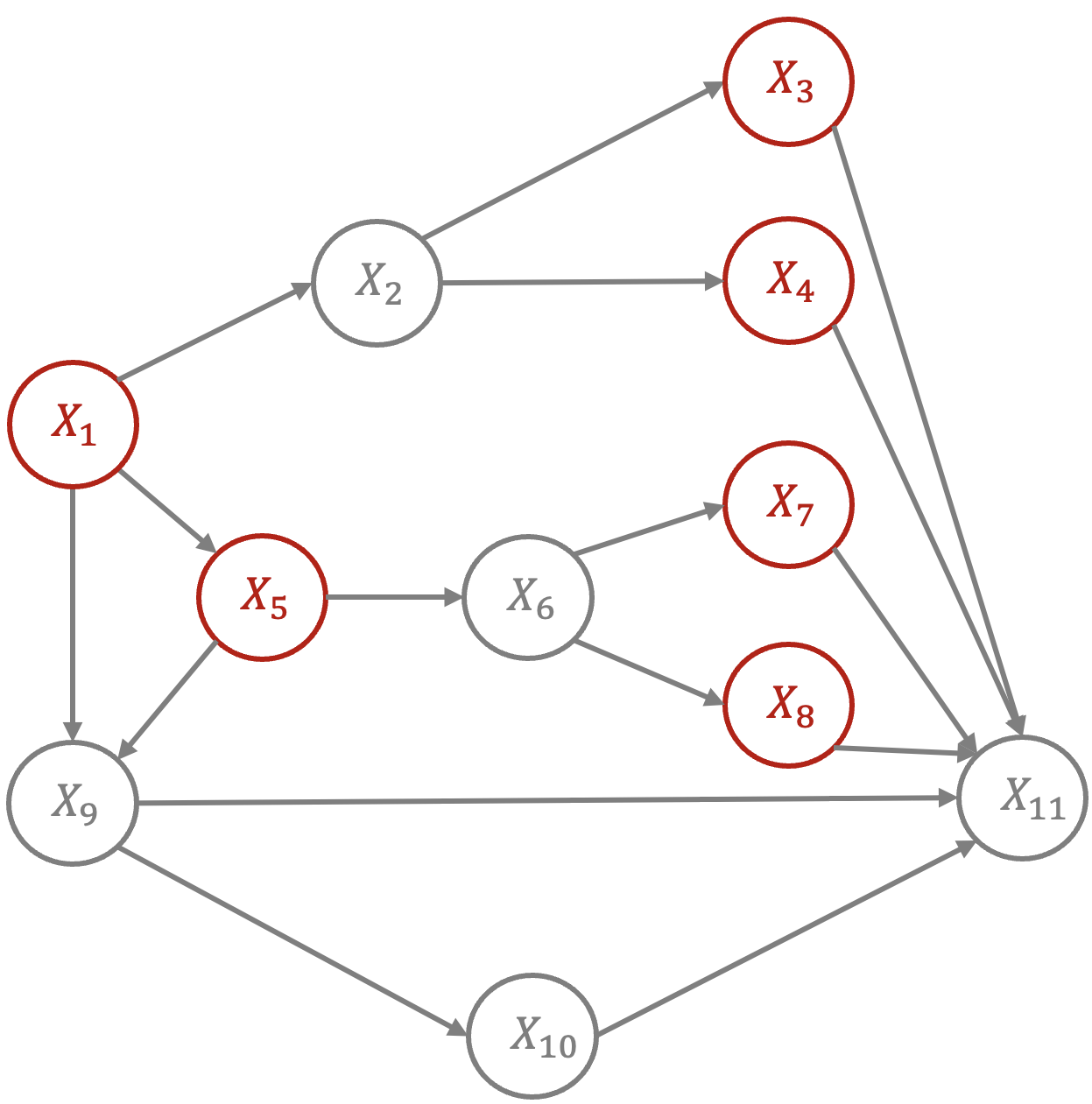}
    \centering
    \caption{SCM $\calM_5$ where the unobserved confounders $X_1$, $X_3$, $X_4$, $X_5$, $X_7$, and $X_8$ exist with five exogenous and endogenous nodes where we intervene in the node $X_9 = x_9$}

    \label{fig_SCM5}
\end{figure}

\begin{example}
We define the set of simple structural equations $f=\{f_i\}_{i \in [11]}$ for SCM $\calM_5$ in Fig. \ref{fig_SCM5} as follows.
\begin{align*}
  X_1 &= f_1(U_1) = U_1 \\
  X_2 &= f_2(X_1, U_2) = -X_1 + U_2 \\
  X_3 & = f_3(X_2, U_3) = X_2 + 0.1 + U_3 \\
  X_4 & = f_4(X_2, U_4) = -X_2 + 0.1 + U_4 \\
  X_5 &= f_5(X_1, U_5) = 1.3 X_1 + X_1 U_5 + U_5 \\
  X_6 &= f_6(X_5, U_6) = -1.2 (X_5 + 0.1) + X_5 + U_6 \\
  X_7 &= f_7(X_6, U_7) = -X_6^2 + X_6 + U_7 \\
  X_8 &= f_8(X_6, U_8) = 3 X_6 + 0.1 + U_8 \\
  X_9 &= f_9(X_1, X_6, U_9) = X_1 X_5 + X_1 - X_5^2 + 0.1 + U_9 \\
  X_{10} &= f_{10}(X_9, U_{10}) = X_9^2 + U_{10} \\
  X_{11} &= f_{11}(X_3, X_4, X_7, X_8, X_9, X_{10}, U_{11}) \\
  &= X_3 X_4 + X_7 X_8 + X_9 X_{10} + X_3 X_9 - X_7 X_{10} - 0.1  \\
\end{align*}
\label{ex_SCM5_simple}
\end{example}

\begin{example}
We define the set of complex structural equations $f=\{f_i\}_{i \in [11]}$ for SCM $\calM_5$ in Fig. \ref{fig_SCM5} as follows.
\begin{align*}
  X_1 &= f_1(U_1) = U_1 \\
  X_2 &= f_2(X_1, U_2) = X_1 (U_2 + 0.1) \\
  X_3 & = f_3(X_2, U_3) = \frac{\sqrt{|X_2|} (|U_3| + 0.1)}{2} + |X_2| + \frac{U_3}{5} \\
  X_4 & = f_4(X_2, U_4) = X_2 + \frac{U_4 + 0.1}{2} X_2 \\
  X_5 &= f_5(X_1, U_5) = -\frac{1}{1 + (|U_5| + 0.1) \exp(-X_1)} \\
  X_6 &= f_6(X_5, U_6) = \frac{U_6 (|X_5| + 0.3)}{5} + U_6 \\
  X_7 &= f_7(X_6, U_7) = X_6 U_7 + |X_6 + 0.01| |U_7| \\
  X_8 &= f_8(X_6, U_8) = 3 X_6 + 0.1 + U_8 \\
  X_9 &= f_9(X_1, X_6, U_9) = X_5^3 X_8 + X_5 + X_8 + U_9 \\
  X_{10} &= f_{10}(X_9, U_{10}) = X_9 U_{10} + (U_{10} + 0.1)^2 \\
  X_{11} &= f_{11}(X_3, X_4, X_7, X_8, X_9, X_{10}, U_{11}) \\
  &= X_3 (X_8 - 0.1) + X_9 X_{10} + X_3 X_9 - X_7 X_{10} + X_3 X_8 \\
  & - X_4 X_9 + X_9 X_{10}  \\
\end{align*}
\label{ex_SCM5_complex}
\end{example}

\section{Additional Result of Synthetic Data Experiment}

\begin{figure}
    \includegraphics[scale=0.335]{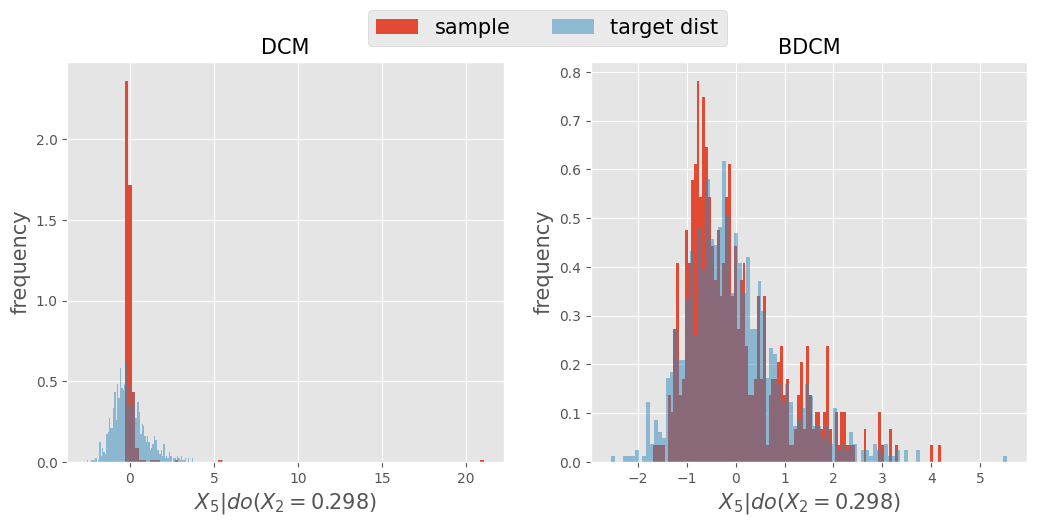}
    \centering
    \caption{Empirical distributions of the $X_5$ sampled from DCM (left) and BDCM (right) compared to the ground-truth target distribution where we intervened in the node $X_2 = 0.298$ in Example \ref{ex_SCM1_complex}}

    \label{fig_res_SCM1_complex}
\end{figure}

\begin{figure}
    \includegraphics[scale=0.335]{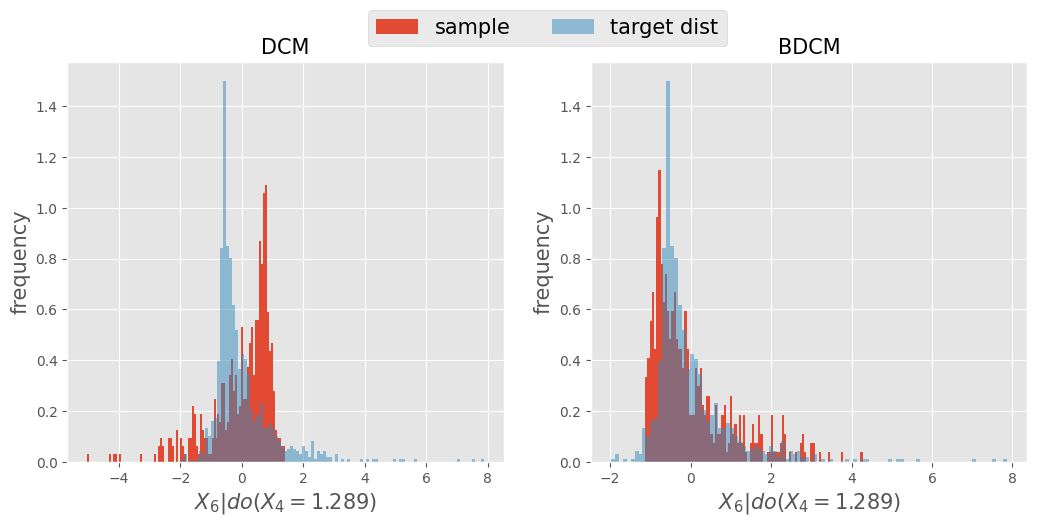}
    \centering
    \caption{Empirical distributions of the $X_6$ sampled from DCM (left) and BDCM (right) compared to the ground-truth target distribution where we intervened in the node $X_4 = 1.289$ in Example \ref{ex_SCM2_complex}}

    \label{fig_res_SCM2_complex}
\end{figure}

\begin{figure}
    \includegraphics[scale=0.335]{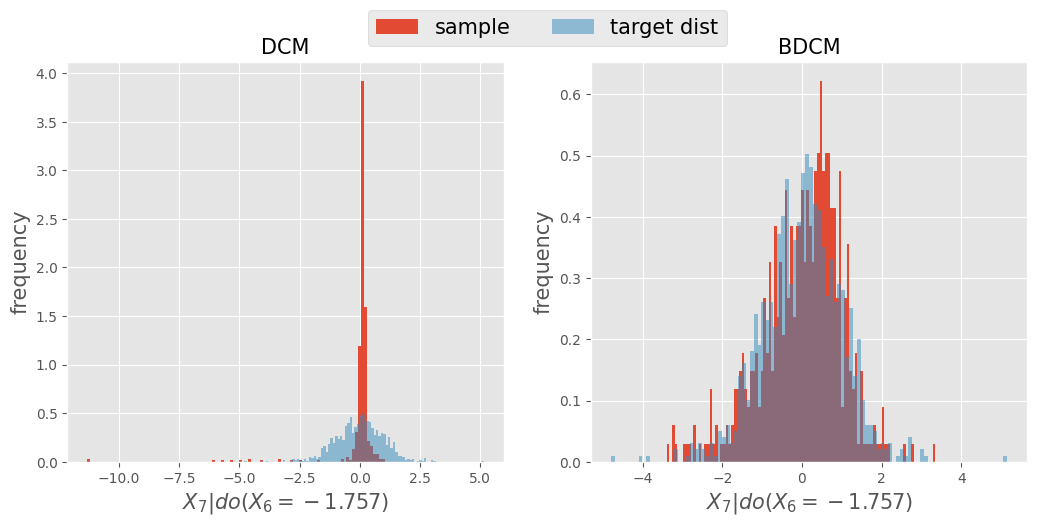}
    \centering
    \caption{Empirical distributions of the $X_7$ sampled from DCM (left) and BDCM (right) compared to the ground-truth target distribution where we intervened in the node $X_6 = -1.757$ in Example \ref{ex_SCM3_simple}}

    \label{fig_res_SCM3_simple}
\end{figure}

\begin{figure}
    \includegraphics[scale=0.335]{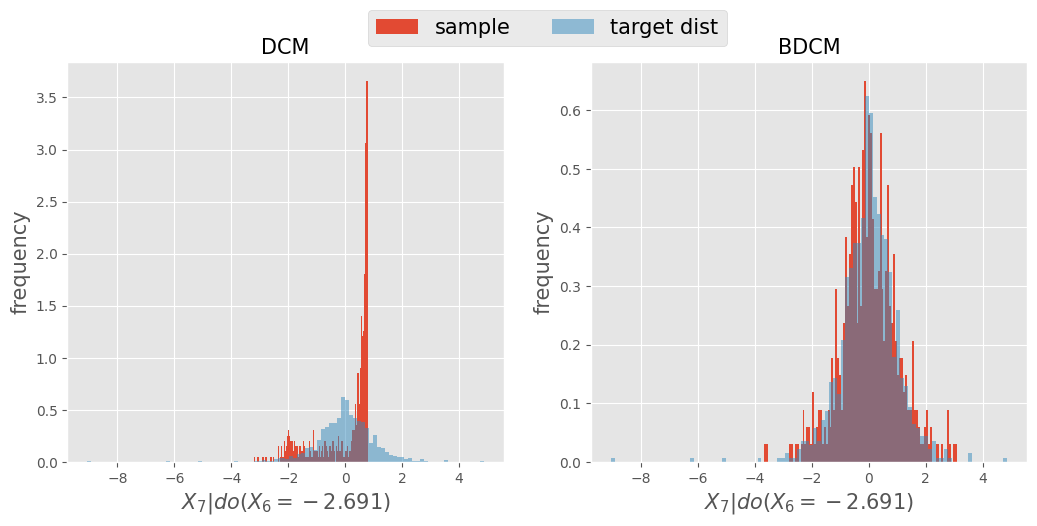}
    \centering
    \caption{Empirical distributions of the $X_7$ sampled from DCM (left) and BDCM (right) compared to the ground-truth target distribution where we intervened in the node $X_6 = -2.691$ in Example \ref{ex_SCM3_complex}}

    \label{fig_res_SCM3_complex}
\end{figure}

\begin{figure}
    \includegraphics[scale=0.335]{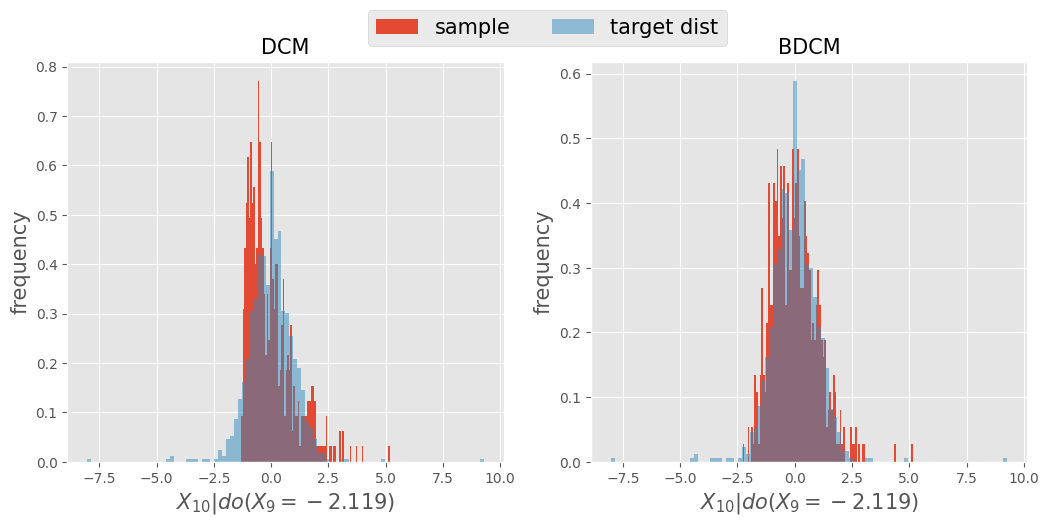}
    \centering
    \caption{Empirical distributions of the $X_{10}$ sampled from DCM (left) and BDCM (right) compared to the ground-truth target distribution where we intervened in the node $X_9 = -2.119$ in Example \ref{ex_SCM4_simple}}

    \label{fig_res_SCM4_simple}
\end{figure}

\begin{figure}
    \includegraphics[scale=0.335]{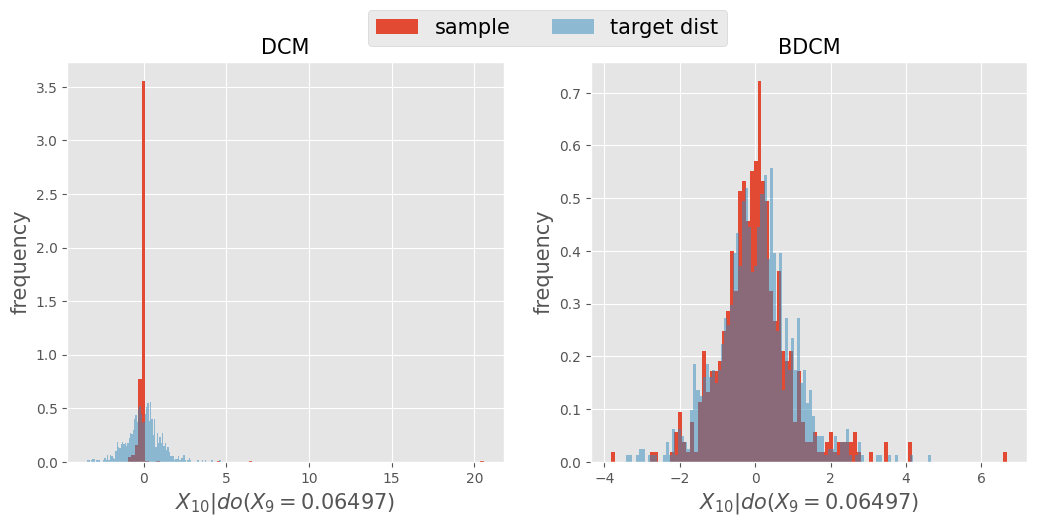}
    \centering
    \caption{Empirical distributions of the $X_{10}$ sampled from DCM (left) and BDCM (right) compared to the ground-truth target distribution where we intervened in the node $X_9 = 0.06497$ in Example \ref{ex_SCM4_complex}}

    \label{fig_res_SCM4_complex}
\end{figure}

\begin{figure}[t]
    \includegraphics[scale=0.335]{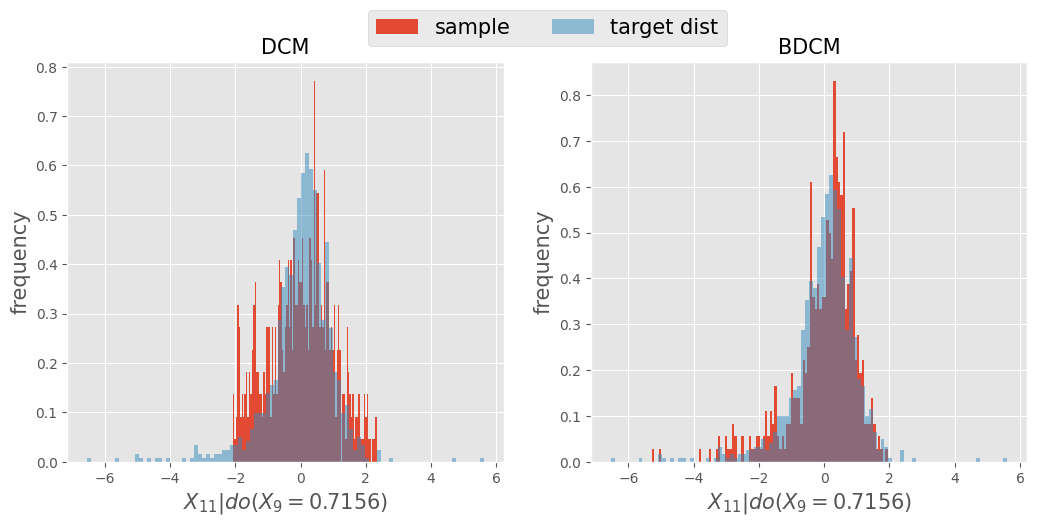}
    \centering
    \caption{Empirical distributions of the $X_{11}$ sampled from DCM (left) and BDCM (right) compared to the ground-truth target distribution where we intervened in the node $X_9 = 0.7156$ in Example \ref{ex_SCM5_simple}}

    \label{fig_res_SCM5_simple}
\end{figure}

\begin{figure}[t]
    \includegraphics[scale=0.335]{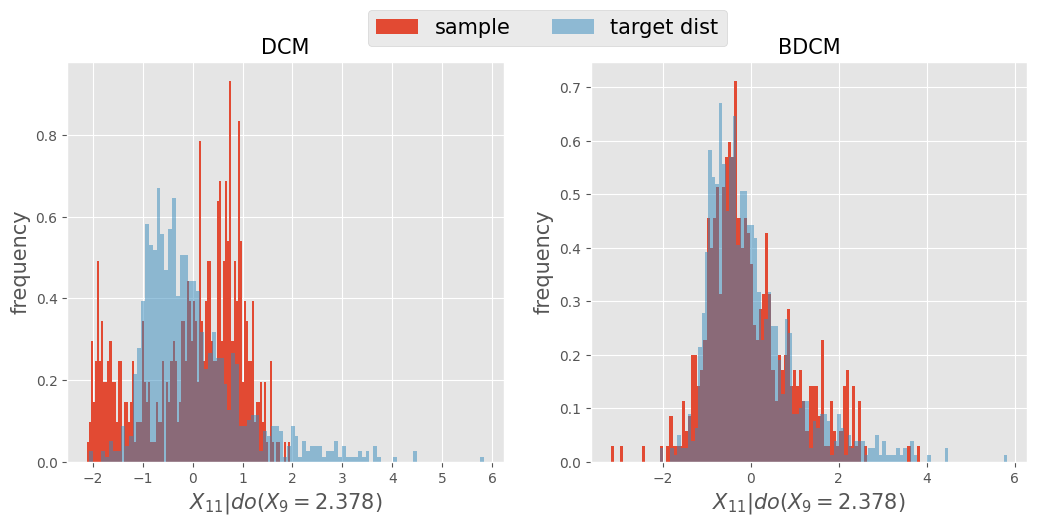}
    \centering
    \caption{Empirical distributions of the $X_{11}$ sampled from DCM (left) and BDCM (right) compared to the ground-truth target distribution where we intervened in the node $X_9 = 2.378$ in Example \ref{ex_SCM5_complex}}

    \label{fig_res_SCM5_complex}
\end{figure}

\end{document}